\documentclass[10pt]{cai}

\usepackage{amsmath,amssymb,amsfonts}
\usepackage{graphicx}
\usepackage{xcolor}
\usepackage[normalem]{ulem}
\graphicspath{{figures/}}

\begin{document}

\volumeheader{35}{0}
\begin{center}

  \title{Discovering Gateway Ports in Maritime Using Temporal Graph Neural Network Port Classification}
  \maketitle

  \thispagestyle{empty}

  \begin{tabular}{cc}
    Dogan Altan\upstairs{\S,*}, Mohammad Etemad\upstairs{\S}, Dusica Marijan\upstairs{\S}, Tetyana Kholodna\upstairs{\affiltwo}
   \\[0.25ex]
   {\small \upstairs{\S} Simula Research Laboratory, Norway} \\
   {\small \upstairs{\affiltwo} Navtor AS, Norway} \\
  \end{tabular}
  
  \emails{
    \upstairs{*}dogan@simula.no
    }
  \vspace*{0.2in}
\end{center}

\begin{abstract}

Vessel navigation is influenced by various factors, such as dynamic environmental factors that change over time or static features such as vessel type or depth of the ocean. These dynamic and static navigational factors impose limitations on vessels, such as long waiting times in regions outside the actual ports, and we call these waiting regions gateway ports. Identifying gateway ports and their associated features such as congestion and available utilities can enhance vessel navigation by planning on fuel optimization or saving time in cargo operation. In this paper, we propose a novel temporal graph neural network (TGNN) based port classification method to enable vessels to discover gateway ports efficiently, thus optimizing their operations. The proposed method processes vessel trajectory data to build dynamic graphs capturing spatio-temporal dependencies between a set of static and dynamic navigational features in the data, and it is evaluated in terms of port classification accuracy on a real-world data set collected from ten vessels operating in Halifax, NS, Canada. The experimental results indicate that our TGNN-based port classification method provides an f-score of 95\% in classifying ports. 

\end{abstract}

\copyrightnotice

\section{Introduction}

\begin{figure*}[!ht]
\begin{center}
     {\includegraphics[scale=0.37]{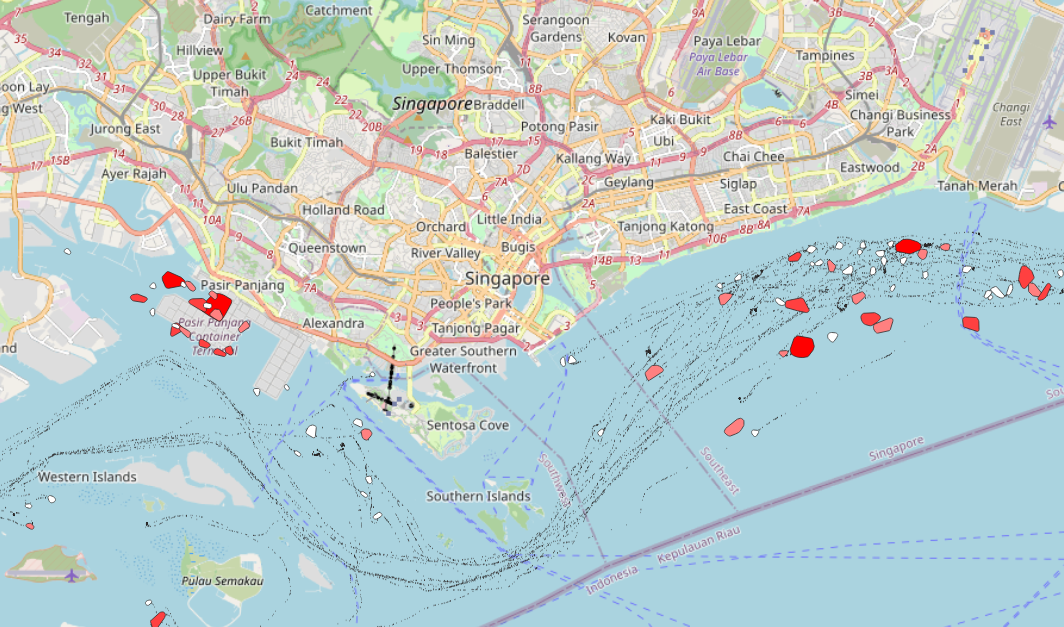}}
\caption{\label{fig:motivation} The illustration of the extracted actual and gateway ports from AIS data based on visiting frequencies. The gateway ports are clustered on the right side with close proximity to actual ports.}
\end{center}
\end{figure*}

According to the United Nations Conference on Trade and Development (UNCTAD) report in 2018, around 80\% of global trade by volume is transferred by vessels and operated cargo activities at ports worldwide \cite{UNC2018}. Therefore, it is crucial to ensure the efficiency of navigation operations to minimize port congestion, reduce vessels' waiting time and optimize travel for reduced vessel fuel consumption. Vessels often need to wait in specific regions outside the actual ports because of port congestion, unfavorable weather conditions, or particular properties of the vessels or the ocean. We call these waiting regions \emph{gateway ports}. To illustrate this type of regions, Figure~\ref{fig:motivation} depicts a port area of Singapore that includes actual and gateway ports. Colored polygons represent the extracted ports, and the extraction is achieved by taking into account the frequency of visits; the darker the color, the higher the number of visits. As can be seen from the figure, gateway ports are mostly clustered outside of the actual ports on the right side of the figure. They can be formed by cargo vessels exchanging items or waiting for better weather conditions. Being able to distinguish between actual and gateway ports allows expert systems to extract the properties of these areas and result in vessels to optimize trajectories for fuel-saving \cite{moon2014impact} and optimize the estimated time of arrival (ETA) \cite{alessandrini2018estimated, park2021vessel}. Furthermore, these types of optimizations contribute to the efficiency of port operations, such as loading/unloading \cite{agra2015maritime} made by port authorities. 

To address the problem of automatically discovering gateways ports, we need to consider spatial relationships between gateway ports since they are generally clustered in close proximity to the actual ports. This type of information is typically obtained by analysing trajectory data collected from multiple vessels over time. Moreover, we need to consider the time dimension, capturing temporal changes of the underlying data, such as waiting times of vessels at ports. Existing solutions to the port classification problem generally focus on analyzing static features of actual ports, lacking a temporal aspect. An effective solution to port classification problem needs to take into account spatio-temporal dependencies of the underlying data. While Graph Neural Networks (GNNs) enable modeling such spatial relations, dynamic graphs enable encoding temporal changes in the graph structure encountered over time, such as the addition or removal of nodes or edges. 

In this paper, we propose a Temporal Graph Neural Network (TGNN) based method for maritime port classification. The graph representation makes it possible to capture underlying spatial relations, and the proposed method processes the information encoded in the graphs temporally to capture temporal dependencies. A time-ordered graph sequence is constructed from the automatic identification system (AIS) data which contains logs obtained from vessels as they sail. The graph is built by partitioning the AIS data into blocks where each block corresponds to a period of time (i.e., daily, weekly). The resulting graph sequence is processed temporally to classify ports. The proposed method is evaluated on real-world AIS data in terms of its effectiveness of correctly classifying ports. The experimental results show that the proposed method classifies ports with an f-score of 95\%. 

The contributions of this study are three-fold:
\begin{itemize}
    \item We propose a novel dynamic graph representation for the port classification problem, capturing spatio-temporal relations between various dynamic and static features of vessel trajectories.
    \item We develop a novel TGNN-based method for discovering gateway ports and classifying target ports as either actual or gateway ports.
    \item We experimentally demonstrate the effectiveness of the proposed method on a real-world data set, showing the classification performance in terms of precision, recall and f-score. 
\end{itemize}

This paper is organized as follows: First, the literature of the described problem is presented. Then, the related notations on the maritime port classification are given. This is followed by the explanation of the proposed port classification method. Afterwards, the evaluation of the proposed method is presented on real-world AIS data. Finally, the paper is concluded with potential future directions.

\section{Related Work}

In this section, we first review existing studies on port classification problem. Then, we briefly present approaches addressing node extraction, which is an auxiliary task in our research.

\subsection{Port classification studies}
Several studies are presented to classify ports considering various port features~\cite{roa2013ports,othman2019sustainable,nguyen2016dry,park2010classification}. Roa et al.~\cite{roa2013ports} classify ports by their sizes and facility types to find out the applicable business models for corresponding types of ports. Another example is Malaysian ports which are classified based on port characteristics such as port size, traffic, and infrastructure~\cite{othman2019sustainable}. Similarly, ~\citet{nguyen2016dry} classify dry ports, which are inland terminals where customers load/unload cargo, considering features such as development reasons, roles, etc.~\citet{park2010classification} present a port classification scheme for container ports considering inland and shipping networks to investigate the economic relationship between ports and their regions. Such existing studies follow qualitative approaches and do not concern with the spatio-temporal data collected from vessels.

Clustering methods are used in the literature to classify ports. Azzam et al.~\cite{azzam2021strategic} present a multi-level k-means clustering approach to cluster and classify ports considering their properties such as harbor type and size. Another work by Mansouri et al.~\cite{mansourimodel} considers k-means clustering to cluster ports taking into account port features such as port size and annual throughput. Such solutions only take into account physical properties of ports while lacking the temporal aspect of the problem.

A recent study by Carlini et al.~\cite{carlini2021understanding} proposes an AIS data representation where the model is adapted from a Global Shipping Network (GSN) structure, with vertices corresponding to ports and edges corresponding to the connections (i.e., voyages) between the ports. In that particular study, the authors show that the movement patterns of some specific types of vessels, such as short-range vessels, change over time due to seasonal trends. Consequently, a temporal analysis of AIS data is crucial. Different from earlier port classification studies, first, we discover gateway ports, which are waiting regions. Second, we classify actual and gateway ports in the maritime domain by considering not only the spatial but also the temporal dependencies of the processed vessel trajectory data. 

\subsection{Node extraction studies}
Vessel navigation movement is accessible in the form of a trajectory which is a series of consecutive spatio-temporal data. Prepossessing such data is required to form the proper inputs for our TGNN model. There are many studies with the focus of finding stop points in a trajectory such as stay point detection~\cite{zheng2011StayPointDetection}, stops and moves of trajectories (SMOT)~\cite{alvares2007model}, clustering-based stops and moves of trajectories (CB-SMOT)~\cite{palma2008clustering}, and direction-based stops and moves of trajectories (DB-SMoT)~\cite{Rocha2010-DB-smot}. These algorithms are categorized as part of trajectory segmentation studies where a trajectory is segmented based on a specific criteria. The utilization of the density-based spatial clustering of applications with noise (DBSCAN) algorithm~\cite{ester1996density} as one of the core algorithms is notable in these studies.

\section{Notations}

This section describes the notations relevant for the maritime port classification problem addressed in the paper.

\paragraph{\textbf{AIS message}} Vessels transmit messages and log their movements during the voyages, which are called AIS messages. Formally, we define each vessel with $s$ and an AIS message is denoted as a tuple, $m = (x,y,t,s,sog)$ where $x$ and $y$ denote GPS coordinates as degrees for longitude and latitude in the EPSG:4326 Geodetic system. $t$ denotes the timestamp of the message, and $sog$ denotes speed over ground, respectively. The entire set of AIS messages is denoted with $M$. 

\paragraph{\textbf{Port}} Vessels navigate in the sea between points, and these points are called ports. A port $p$ is represented as a tuple $(id,x,y)$ where $id$ corresponds to the unique identifier of the port and $x$ and $y$ denote the center of the GPS coordinates as degrees for longitude and latitude in the EPSG:4326 Geodetic system. We also consider a spatial function $\gamma(p, c)$, which specifies a polygon as the port region with an ordered sequence of point sequences $c$ that defines the corresponding polygon, and it is centered on coordinates of the specific port $p$. The global set of ports are denoted as $p \in P$.

\paragraph{\textbf{Trajectory}} Vessels follow a series of points (coordinates) in the sea and these time-ordered consecutive coordinates make up the trajectory of that particular vessel. In this study, we represent a trajectory of a vessel $s$ with ${tr}_s= ((x,y)_0,(x,y)_{1},...,(x,y)_n)$ where $(x,y)$ is a binary tuple that includes degrees for longitude and latitude.

\paragraph{\textbf{Visit}} Vessels navigate and visit ports to fulfill their business objectives. We explicitly define a visit $v = (p, m, t)$ of the vessel $s$ to a port $p$ at time $t$ with the condition that there is at least one $m \in M$ with coordinates $x$ and $y$ in the port area given by $\gamma(p, c)$ considering the ordered points $c$ that define the port.

\paragraph{\textbf{Voyage}} Two consecutive visits to different ports are considered voyages, and it is denoted as a pair of visits $voy = (v_1, v_2)$ and the same vessel performed both visits ($v_1(s) = v_2(s)$). One should also note that these visits need to be consecutive in terms of time ($v_1(t) < v_2(t)$), and there should be no visits to other ports performed by the corresponding vessel between $v_1(t)$ and  $v_2(t)$. The ports of $v_1$ and $v_2$ satisfying these criterion are called origin and destination ports, respectively.

\section{TGGN-based Port Classification Approach }

Our proposed approach to maritime port classification is based on Temporal Graph Neural Networks (TGNNs). In this section, we first present preliminaries of the TGNN structure, then elaborate on the proposed port classification approach.

\subsection{Temporal Graph Neural Networks (TGNNs)}

A graph neural network (GNN) is represented with $G = (V,E)$ where $V$ is the vertex set and $E$ is the edge set of the graph G. In a GNN,  information is embedded into nodes and edges to come up with inferences. Each node is denoted with $h_i \in V$, and each edge is denoted with $e_{ij}$ where $i$ and $j$ denote node indices. Consecutive graphs ordered by time make up a TGNN, and it includes layers, each of which is denoted with $l$ corresponding to a distinct time step. Information is encoded into the nodes and edges in this temporal graphs. Node encodings correspond to $enc(h_i^l)$ where $l$ is the layer index and $i$ is the node index in the corresponding graph layer. Edge encodings correspond to $enc(e_{ij}^l)$ where $l$ is the layer index and $i$ and $j$ denote the corresponding node indices. Node-level, edge-level and graph-level inferences are possible by taking into account node and edge encodings. In this particular study, we adopt a node-level prediction method to classify ports. The upcoming section presents the details of how we adapt this formulation for our TGNN-based method.

\subsection{The Proposed Algorithm}

The dynamic graphs underlying our TGNN consist of nodes ($V$) as ports ($P$). A node encoding at the layer $l$ ($enc(h_i^l)$) corresponds to the features tailored from port information and the AIS message set ($M$). An edge encoding at the layer $l$ ($enc(e_{ij}^l)$) corresponds to the distances  between the corresponding ports ($\delta(p_i, p_j)$ where $\delta$ is haversine distance function and $i$ and $j$ are port indices).  
        
Our proposed TGNN-based method includes three stages: data preprocessing, model training and port classification. Tasks such as node extraction, segment annotation, voyage extraction and temporal graph construction are performed in the data preprocessing stage. This is followed by the model training stage where temporal graphs are used to train a TGNN model. The final stage is port classification, and
the trained model is used to classify ports with the labels \emph{actual} and \emph{gateway} ports.

\begin{figure}[ht]
\begin{center}
     {\includegraphics[scale=0.42]{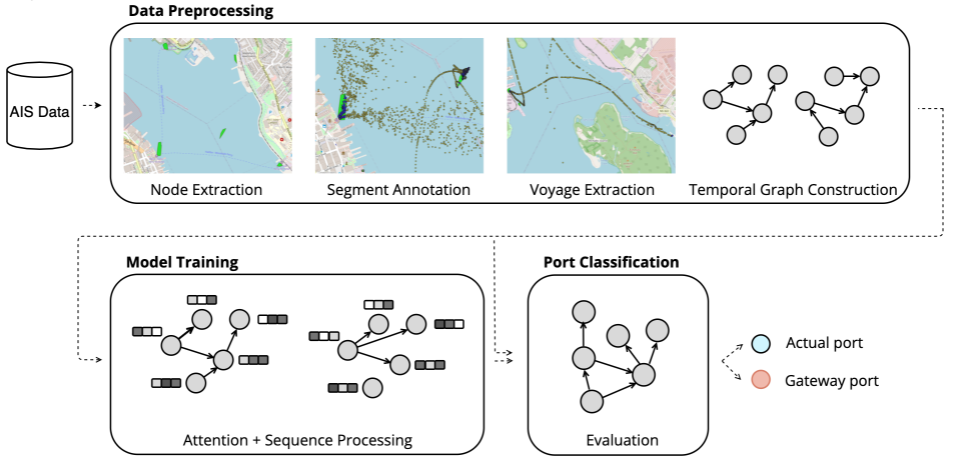}}
\caption{\label{fig:tgn-overall} The stages of the proposed port classification algorithm.}
\end{center}
\end{figure}

Figure~\ref{fig:tgn-overall} illustrates the stages of the proposed port classification method. First, ports are extracted from the log messages obtained from vessels, AIS data (node extraction). Then, each AIS message is annotated, whether it is a message obtained while the vessel was within a port region or on the sea (segment annotation). Later, voyages of vessels are extracted (i.e., the start and end ports of the vessels) in the voyage extraction stage. Afterwards, the input data is represented as a sequence of graphs (temporal graph construction). Finally, a model is trained for the port classification task, and it is evaluated. The upcoming subsections elaborate the details of these tasks.

\paragraph{\textbf{Node Extraction}} 
The aim of the node extraction task is to extract ports from the AIS data. We use DBSCAN algorithm to extract ports. DBSCAN has two parameters: epsilon ($eps$) and the minimum number of points in each cluster ($minPts$). We experiment with different values for these parameters, and our experiments show that using $minPts=20$ and $eps=0.005$ degree which is near to 500 meters is a reasonable decision and can identify nodes that are more relevant to a subject matter expert. We use spatial operation and PostGIS to keep a set of nodes and merge newly discovered ports using a buffered convex-hull of the adjacent port and the newly discovered port. 

\paragraph{\textbf{Segment Annotation}}

After the node extraction task, the segment annotation task takes place. This task aims to label all AIS messages $m \in M$ that include trajectory points $(x,y)$ with the following labels: \textit{id} or \textit{seapoint}. The label \textit{id} indicates that the corresponding trajectory point falls into a port region that is extracted in the previously described node extraction task, and its label is the id of that particular port. If a trajectory point is not located within any of the extracted ports, then it is annotated with the label \textit{seapoint}. Formula \ref{formula:annotation} formulates this segmentation task as a partial function.

\begin{equation}
label(m_i) = \begin{cases} 
    p_j.id & \exists p_j \; (m_i.(x,y) \in \gamma(p_j,c_j))\\
      \textit{seapoint} & otherwise
   \end{cases}
   \label{formula:annotation}
\end{equation}

\paragraph{\textbf{Voyage Extraction}}

Voyage extraction task requires investigating the annotated trajectory points to extract voyages. This necessitates keeping track of the ordering of segment annotations. To exemplify, once an AIS message that contains an \textit{id} as its label is encountered, the corresponding port with that \textit{id} is considered as the source of the voyage ($v_i$). Then, this voyage is tracked until another AIS message with an \textit{id} as its label is encountered after a number of sequential \textit{seapoint} labels. This port is then considered as the destination of the corresponding voyage ($v_j$), and the voyage $voy = (v_i, v_j)$ is registered as a new voyage.

\paragraph{\textbf{Temporal Graph Construction}}

After the voyages are extracted from the AIS data, the temporal graph construction task takes place. For each pre-defined time interval, a graph $G^l$ is generated with the information encoded on it. In this study, we take the time interval a single calendar day and generate daily graphs. We empirically set this interval to a single calendar day as the used data set for the evaluation includes AIS messages of passenger vessels operating between ports located within proximity. Port-related information is encoded in the nodes ($enc(h_i^l)$) for the corresponding day as we associate ports with nodes in our representation. The following features extracted from AIS messages are embedded in the nodes: port visit frequencies, waiting times in the ports, and port arrival average speed statistics. This information is concatenated to form a vector that represents node features for each node. Considering the taxonomy presented by Rozemberczki et al.~\cite{rozemberczki2021pytorch} to represent spatiotemporal input data, our design falls into the category \textit{dynamic graph with temporal signal} as both the graph structure (i.e., addition/removal of nodes/edges) and the node features (i.e., the embedded information on nodes) change at each time step. The following items elaborate on each tailored feature.

\begin{itemize}
    \item Port visiting statistics: This feature corresponds to the visiting statistics of ports by the vessels within a period of time. This information is extracted for each port taking into account the number of arrivals and departures of vessels to and from that particular port. This statistic is calculated separately as a voyage may start and end on different calendar days. 
    \item Waiting time: This feature corresponds to the vessels' existence times in port regions for a given period of time in terms of minutes.
    \item Speed statistics: This feature corresponds to the average speeds of vessels during the voyages. We extract this feature for each port. We also consider average speeds of vessels while they are in a port region.
\end{itemize}

As edges are associated with the connectivity of ports (i.e., there is at least one voyage between the ports $p_i$ and $p_j$ occurred on a day corresponding to the layer $l$ on the graph), we utilize the haversine distance between the nodes as edge encodings ($enc(e_{ij}^l)$). At the end of the temporal graph generation task, a time-ordered temporal graph sequence is obtained with the node embeddings as port-related features, and edges as the adjacency relations (i.e., the distance between ports).

\begin{figure}[h]
\begin{center}
     {\includegraphics[scale=0.25]{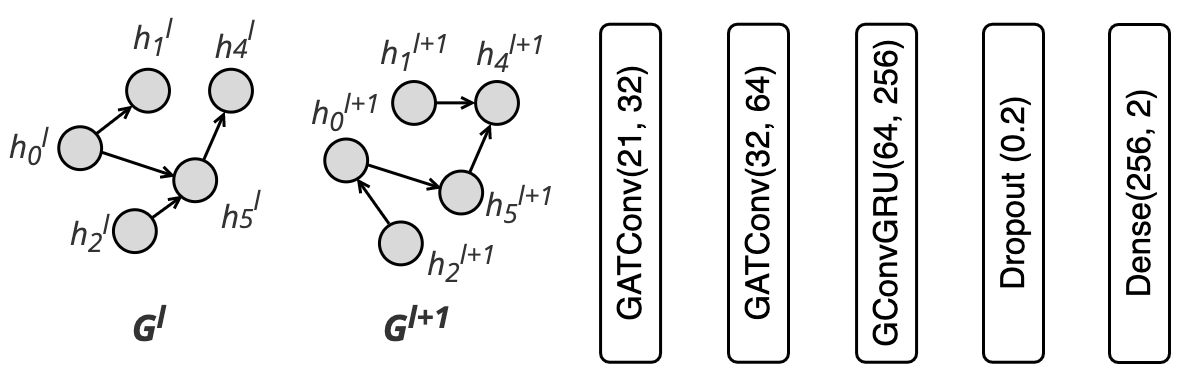}}
\caption{\label{fig:tgn-desing} The designed TGNN for the port classification problem.}
\end{center}
\end{figure}

To learn and capture underlying patterns in the AIS data, we use a TGNN design which is depicted in Figure~\ref{fig:tgn-desing}. We employ two attention-enabled graph convolution layers, namely GATConv \cite{velivckovic2017graph}. Each convolution layer includes two attention heads with a mean aggregation scheme for message passing, and the edge dimension is set to one. These convolution layers are followed by a recurrent graph convolution layer that builds on gated recurrent units (GRUs), namely, GConvGRU~\cite{seo2018structured}. The graph sequence is processed in this layer and the result is fed into a dropout layer. Later on, a dense layer is employed which is followed by a softmax layer to map the input to classes.

\section{Experimental Evaluation}

This section presents the experimental setup followed by the experimental results answering the three research questions. We evaluate the proposed method in terms of its capability and robustness on a real-world dataset. Specifically, we are interested in answering the following research questions:
\begin{enumerate}
    
    \item [\textbf{RQ1}] Is the model effective in classifying maritime ports?
    
    \item [\textbf{RQ2}] How do the spatio-temporal relations affect the port classification performance?
      
    \item [\textbf{RQ3}] Is the model robust under different configuration changes, i.e., different dropout probabilities as a regularization parameter?
\end{enumerate}

\subsection{Experimental Setup}

\begin{figure}[!h]
\begin{center}
     {\includegraphics[scale=0.4]{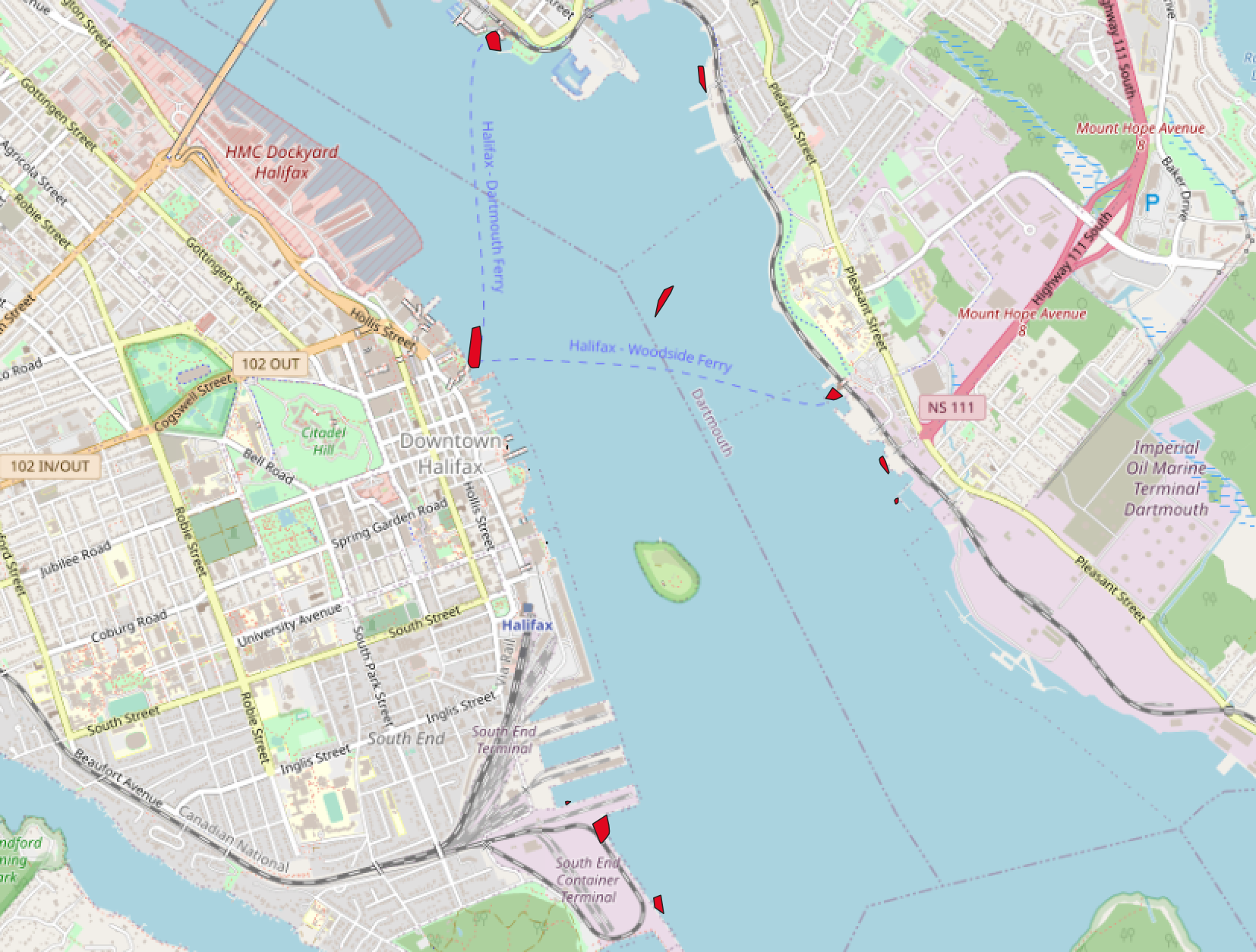}}
\caption{\label{fig:halifax-ports} The port region of Halifax used in the experiments. Red regions denote the extracted ports.}
\end{center}
\end{figure}

\textbf{Dataset.} 
We use a publicly available\footnote{https://github.com/metemaad/WS-II} real-world data set to evaluate our presented port classification method. The dataset includes 513012 AIS messages obtained from 10 vessels operating within a port region of Halifax, NS, Canada between March 2019 and July 2019, shown in Figure~\ref{fig:halifax-ports}. There is a total of 15 ports extracted, and for the sake of space, only 10 ports are shown in the figure.

\textbf{Evaluation metrics.} We first evaluate the performance of the proposed method in terms of its effectiveness in accurately classifying ports by presenting confusion matrices. Then, we analyze the classification performance with respect to the following classification scores: precision, recall and f-score. Finally, we evaluate changes of training and validation losses during the training.

\textbf{Training.} We train our model to learn a total number of 500066 trainable parameters. We use time series split cross validation to validate our model, and we partition the data by taking into account the time dimension of the input graph sequence. We set the number of split as eight. An early stopping scheme is applied to decide on when to stop training. It considers the validation loss with an empirically set patience threshold. To tackle the imbalanced class distribution in the data set, we use adaptive weights while updating the model's parameters during training; that is, the more instance a class has, the less weight it has. We utilize PyTorch for model implementation and PyTorch Geometric Temporal library \cite{rozemberczki2021pytorch} for processing spatio-temporal data. Table~\ref{param-table} lists the parameters that are used during training.

\begin{table}[h]
\centering
\caption{\label{param-table}Hyperparameters that are used during training phase.}
\resizebox{5.5cm}{!}{%
\begin{tabular}{cc} 
Parameter        & Value \\ \hline
Optimizer        &   Adam    \\ \hline
Loss             &  Cross Entropy     \\ \hline
Learning rate ($\eta$) &  $10^{-4}$     \\ \hline
Dropout &  $0.2$      \\ \hline
Patience         &     10 
\end{tabular}
}
\end{table}
\subsection{Experimental Results and Analysis}

\subsubsection{Classification Effectiveness}
\begin{figure*}[h]
     \centering
     \begin{subfigure}[b]{0.3\textwidth}
         \centering
         \includegraphics[width=\textwidth]{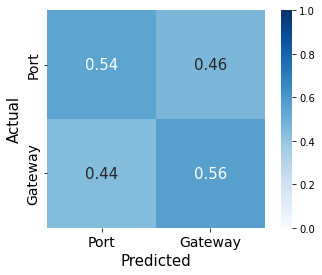}
         \caption{}
         \label{fig:cm-wo-temporal}
     \end{subfigure}
     \hfill
     \begin{subfigure}[b]{0.3\textwidth}
         \centering
         \includegraphics[width=\textwidth]{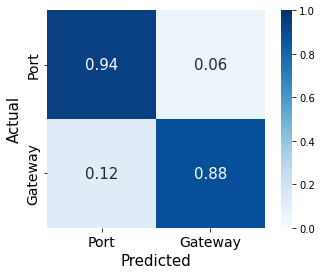}
         \caption{}
         \label{fig:cm-wo-att}
     \end{subfigure}
     \hfill
     \begin{subfigure}[b]{0.3\textwidth}
         \centering
         \includegraphics[width=\textwidth]{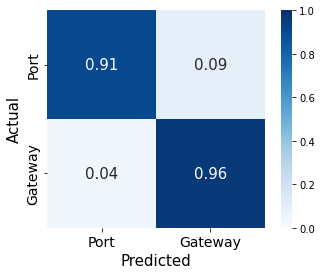}
         \caption{}
         \label{fig:cm-w-temporal}
     \end{subfigure}

        \caption{The normalized confusion matrices for (a) the TGNN method w/o temporal analysis (b) w/o attention (c) w/ both temporal analysis and attention.}
        \label{fig:conf-matrices}
\end{figure*}

Figure~\ref{fig:conf-matrices} presents the normalized confusion matrices of the proposed method for the different settings. Vertical axes denote the actual labels, and horizontal axes denote the predicted labels by the method. The first confusion matrix (Figure~\ref{fig:cm-wo-temporal}) depicts the classification performance of the proposed method where a temporal analysis on the input graphs is excluded (i.e., the graphs are processed individually without any propagation of temporal information). The second confusion matrix (Figure~\ref{fig:cm-wo-att}) depicts the results of the setting where the attention-enabled graph convolution layers are excluded in the design. The last confusion matrix (Figure~\ref{fig:cm-w-temporal}) depicts the obtained results with the proposed method where consecutive input graphs are processed, taking into account the temporal dimension, and activating the attention-enabled graph convolution layers. 

As can be seen from Figure~\ref{fig:cm-wo-temporal}, when temporal analysis is excluded, the method predicts only 54\% of the ports accurately, while it predicts 56\% of the gateway ports accurately. 
Considering the results in Figure~\ref{fig:cm-wo-att}, excluding only attention yields to predicting 94\% of the ports accurately. This setting also provides an improvement in predicting gateway ports with a score of 88\%. Incorporating attention and temporality together provides a classification score of 91\% for ports, which is less than 3\% compared to the previous setting. On the other hand, this setting provides a 96\% classification score for the gateway ports, which is 8\% more than the setting where the attention mechanism is excluded.

\begin{table}[h]
\centering
\caption{Performance analysis of the proposed method with different settings.\label{table:ablation}}
\resizebox{8cm}{!}{%
\begin{tabular}{ccccc}
\cline{3-5}
 &                  & \multicolumn{1}{c}{Precision} & \multicolumn{1}{c}{Recall} & \multicolumn{1}{c}{F-Score} \\
\multicolumn{1}{c}{$\alpha$}  &  \textit{t} & $\mu \pm \sigma$ & $\mu \pm \sigma$  & $\mu \pm \sigma$   \\ \hline
  $\checkmark$  &  &  0.90 $\pm$ 0.02  & 0.55 $\pm$ 0.16  & 0.59 $\pm$  0.14   \\ \hline
   & $\checkmark$  &  0.97 $\pm$ 0.04  & 0.93 $\pm$ 0.07  & 0.94 $\pm$  0.06   \\ \hline

  $\checkmark$    & $\checkmark$  & 0.98 $\pm$ 0.02 &   0.93 $\pm$ 0.05   &  0.95 $\pm$  0.06                       
\end{tabular}
}
\end{table}

\subsubsection{Effect of Spatio-temporal Relationships}
Table~\ref{table:ablation} presents the classification scores in terms of precision, recall and f-score. Each row presents the weighted scores of a distinct setting, and each column presents the score for the corresponding metric. $\alpha$ denotes whether attention is used, and $t$ stands for whether a temporal analysis is applied to the input. As can be seen from the table, the setting where only attention is used provides an f-score of 59\%. On the other hand, the setting where only temporality is used provides an f-score of 94\%, which is significantly better than the previous setting where only attention is used. When attention and temporality are incorporated, a slight performance improvement is achieved with an f-score of 95\%. In our representation, temporality enables the graph sequence to propagate information between graph instances through time. Attention enables the nodes that correspond to ports to be aware of their neighbor ports by considering their attention scores. Considering the misclassified instances, those settings where attention or temporality is excluded confuse labels most of the time when the graph constructed for that corresponding day is sparse (i.e., there are isolated nodes in the graph which means some ports are not visited during that day). In such graphs, the spatio-temporal relationships are fewer. 

\subsubsection{Classification Robustness to Configuration Changes}
To avoid overfitting and make the model generalize better, we use dropout and early stopping as regularization techniques. Table~\ref{table:train-analysis} presents the effect of the dropout probability choice on the performance. Each row presents the results of a different dropout probability setting, and each column presents the scores of a distinct metric. First, we analyze the case where the dropout layer is disabled. This setting provides an f-score of 0.89 as a baseline. When the dropout probability is set to 0.1, an f-score of 0.94 is achieved. Setting the dropout probability to 0.2 provides the best result with an f-score of 0.95. As the dropout probability is set to 0.3, the obtained f-score drops to 0.92. In our experiments, we choose the setting where the dropout probability is 0.2 as it provides the best f-score.

\begin{table}[h]
\centering
\caption{Performance analysis of the proposed method for different dropout probabilities.\label{table:train-analysis}}
\resizebox{8cm}{!}{%
\begin{tabular}{cccc}
\cline{2-4}
  & \multicolumn{1}{c}{Precision} & \multicolumn{1}{c}{Recall} & \multicolumn{1}{c}{F-Score} \\
\multicolumn{1}{c}{Probability}  & $\mu \pm \sigma$ & $\mu \pm \sigma$  & $\mu \pm \sigma$   \\ \hline
0 &  0.97 $\pm$ 0.04  & 0.86 $\pm$ 0.07  & 0.89 $\pm$  0.05   \\ \hline
0.1 & 0.97 $\pm$ 0.02 &   0.91 $\pm$ 0.09   &  0.94 $\pm$  0.07  \\ \hline
0.2 & 0.98 $\pm$ 0.02 &   0.93 $\pm$ 0.05   &  0.95 $\pm$  0.06  \\
\hline
0.3 & 0.97 $\pm$ 0.02 &   0.91 $\pm$ 0.08   &  0.92 $\pm$  0.06 \\
\end{tabular}
}
\end{table}

\begin{figure}[h]
\begin{center}
     {\includegraphics[scale=0.7]{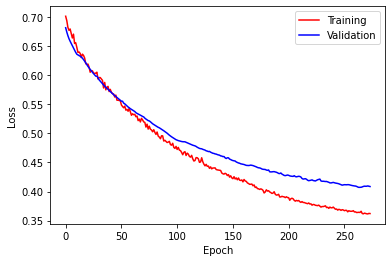}}
\caption{\label{fig:loss} Training and validation loss. Early stopping scheme stops training when the validation loss starts to increase.}
\end{center}
\end{figure}

Figure~\ref{fig:loss} presents an analysis on training and validation loss during the  model training phase. The y-axis denotes the loss, and the x-axis stands for the epoch number. The red line indicates the training loss and the blue line indicates the validation loss. In this particular training instance, the validation loss starts to increase around epoch 270. At this point, we stop the training and select the target model for evaluation.

\section{Discussion and Conclusion}
This research addresses a significant problem in the maritime domain, which is the port classification problem. It deals with discovering gateway ports (which are not actual ports but waiting regions for vessels) and distinguishing between gateway and actual ports. To this end, we present a novel TGNN-based port classification method that processes spatio-temporal AIS data. The proposed method first extracts ports from the raw AIS data then annotates segments in the AIS messages. Later, the voyages are extracted, taking into account the extracted port information and the segment annotations. This is followed by constructing a time-ordered daily graph sequence encoded with these extracted features. The performance of the proposed method is evaluated on a real-world data set consisting of logs collected from vessels sailing in Halifax, NS, Canada, in terms of precision, recall and f-score. The conducted experiments show that our TGNN-based port classification method has the ability to effectively classify maritime ports by taking into account spatio-temporal dependencies of the underlying data with an f-score of 95\%. Furthermore, the experiments demonstrate that excluding attention (which evaluates the importance of neighboring ports as a spatial analysis) or temporal analysis from the proposed method yields degraded performance. 

We believe that this study fills a gap in the literature by bringing up the discovery of gateway ports and distinguishing them from actual ports. The preliminary obtained results are promising in terms of the achieved classification scores on the available real-world AIS data set. We are aware of the fact that the data set used in the experiments is limited in terms of the number of considered ports. It is worth to note that one of the main challenges of the port classification problem is the lack of available benchmark data sets annotated with the ground truth. Therefore, we believe this study presents an important initial step for the discovery of gateway ports and classifying gateway and actual ports. We leave the investigation of the port classification problem with a more extensive number of ports located in different regions and a more comprehensive comparative analysis with other baselines as future work. We also plan to process the detected gateway port information to optimize vessel trajectories by applying graph optimization algorithms.

\section*{Acknowledgements}
This work is supported by ECSEL JU under grant agreement No 101007260 and the Research Council of Norway under grant agreement No 329090.


\printbibliography[heading=subbibintoc]
\end{document}